# Comprehensive Study on Sentiment Analysis: From Rule based to modern LLM based system

Shailja Gupta[1]     Rajesh Ranjan[2]     Surya Narayan Singh[3]

**Abstract**

This paper provides a comprehensive survey of sentiment analysis within the context of artificial intelligence (AI) and large language models (LLMs). Sentiment analysis, a critical aspect of natural language processing (NLP), has evolved significantly from traditional rule-based methods to advanced deep learning techniques. This study examines the historical development of sentiment analysis, highlighting the transition from lexicon-based and pattern-based approaches to more sophisticated machine learning and deep learning models. Key challenges are discussed, including handling bilingual texts, detecting sarcasm, and addressing biases. The paper reviews state-of-the-art approaches, identifies emerging trends, and outlines future research directions to advance the field. By synthesizing current methodologies and exploring future opportunities, this survey aims to understand sentiment analysis in the AI and LLM context thoroughly.

## 1. Introduction

### 1.1. Background and Motivation

Sentiment analysis, also known as opinion mining, has become a pivotal area of research in natural language processing (NLP) and artificial intelligence (AI). This field focuses on identifying and extracting subjective information from textual data, often to determine the sentiment expressed by individuals or groups. The rapid growth of digital communication platforms has amplified the need for effective sentiment analysis tools to manage and interpret vast amounts of user-generated content. Traditional methods of sentiment analysis, based on manually crafted rules and lexicons, have been increasingly supplemented by advanced machine learning and deep learning approaches, which offer more nuanced and scalable solutions (Liu et. al. 2012) (Zhang et. al. 2015).

---

[1] Product Manager, New Jersey, USA
[2] Product Manager, California, USA
[3] Software Developer, Hyderabad, India

## 1.2. Objectives of the Survey

The primary objective of this survey is to provide a comprehensive overview of sentiment analysis techniques, from traditional methods to cutting-edge deep learning models. This paper aims to:

- Review the evolution of sentiment analysis techniques and their impact on the field.
- Highlight key challenges and limitations associated with sentiment analysis, including issues related to multilingual texts and sarcasm detection.
- Explore the role of large language models (LLMs) in advancing sentiment analysis capabilities.
- Identify emerging trends and suggest future research directions to address existing challenges and enhance the effectiveness of sentiment analysis.

## 2. Sentiment Analysis: Fundamental Concepts

## 2.1. Definition and Scope

Sentiment analysis is a subfield of natural language processing (NLP) that focuses on determining and analyzing the emotional tone expressed in text. It involves classifying text data such as positive, negative, or neutral, or more granular emotional states such as joy, anger, or sadness (Pang et. al. 2008). This process helps in understanding public opinion, customer satisfaction, and various emotional insights from textual data.

## 2.2. Historical Development

The development of sentiment analysis can be traced back to the early 2000s when initial approaches were primarily rule-based. Early systems relied on handcrafted lexicons and pattern matching to detect sentiments (Turney et. al. 2003). With the advancement of machine learning, these methods evolved to include statistical models and feature-based classifiers. The introduction of deep learning and neural networks in the 2010s marked a significant shift, allowing for more sophisticated and accurate sentiment analysis by leveraging vast amounts of data and computational power (Kim et. al. 2014).

## 2.3. Key Challenges and Objectives

Sentiment analysis, a crucial area in natural language processing (NLP), aims to detect and interpret emotions expressed in text. However, several challenges hinder its accuracy and effectiveness. As sentiment analysis has evolved, the complexity of linguistic nuances, cultural variations, and context-dependent meanings have posed significant obstacles. This section discusses three key challenges: ambiguity and context, sarcasm and irony, and multilingual or bilingual texts, all of which complicate the accurate interpretation of sentiment.

### 1. Ambiguity and Context

Ambiguity in language is one of the foremost challenges in sentiment analysis. Words and phrases often carry multiple meanings depending on their context, which makes determining sentiment a complex task. For instance, consider the word "cold." In isolation, this word could indicate a physical temperature or describe someone's distant personality. In another context, it might imply negative emotions like rejection or loneliness. This contextual dependency creates a significant hurdle for traditional rule-based models, which may struggle to accurately interpret nuanced meanings. Even advanced deep learning models like BERT and GPT, which are designed to handle context better by considering the surrounding words, face limitations when processing ambiguous expressions. The challenge of understanding context is particularly relevant in fields like product reviews, where a word like "light" might refer to the weight of a product in one review and to its brightness in another. This polysemy often results in inaccurate sentiment classification if the system fails to understand the broader context. Current research focuses on improving context-awareness through advanced architectures such as transformers, but this remains an open area of exploration in NLP.

### 2. Sarcasm and Irony

Detecting sarcasm and irony is another formidable challenge in sentiment analysis. These linguistic constructs rely on the discrepancy between literal and intended meanings, which are often difficult to detect using traditional or even some modern machine learning methods. For instance, the phrase "Oh, great! Another delay!" is sarcastic, with an underlying negative sentiment, despite the use of a seemingly positive word, "great." Sarcasm and irony are particularly challenging because they require the system to not only understand the literal text but also infer the speaker's tone, context, and intent. This task demands a deeper understanding of pragmatics, which goes beyond the capabilities of simple keyword-based or lexicon-based approaches. Researchers have started using multimodal approaches, incorporating not just text but also voice, facial expressions, or other signals, to improve the detection of sarcasm and irony. However, these techniques are still under development and often lack generalizability across different domains.

3. **Multilingual and Bilingual Texts**

Analyzing multilingual and bilingual texts introduces further complexity to sentiment analysis. Sentiment can vary dramatically between languages due to differences in grammar, idioms, and cultural expressions. For example, sentiment conveyed in English may not have a direct equivalent in languages like Chinese or Arabic, resulting in a potential loss of meaning during translation. Moreover, sentiment analysis systems trained in one language often fail to generalize to another, primarily due to the lack of annotated data in many languages. Multilingual models, such as mBERT, have attempted to address this issue by training in a variety of languages simultaneously, but they still face problems when dealing with code-switching (i.e., alternating between languages within the same sentence). Additionally, cultural differences in expressing emotions can further complicate the analysis. For example, while direct expressions of dissatisfaction might be common in Western languages, indirect or more formal expressions are preferred in some Asian languages. This disparity makes sentiment detection harder in multilingual contexts. Handling bilingual texts introduces the additional challenge of translation accuracy. Automated translation tools are often employed, but their limitations can distort the sentiment. A misinterpretation of a single word can flip the overall sentiment of a sentence, thus reducing the reliability of sentiment analysis in multilingual settings.

The objectives of sentiment analysis are to:

- Extract and quantify subjective information from text.
- Provide actionable insights for businesses, policymakers, and researchers.
- Enhance understanding of emotional trends and public opinion.

**2.4. Types of Sentiment Analysis**

Sentiment analysis can be categorized into several types based on the granularity and focus of the analysis:

**Document-Level Sentiment Analysis**: Classifies the sentiment of entire documents or reviews, often used for summarizing overall opinion (Hu et. al. 2004).

**Sentence-Level Sentiment Analysis**: Focuses on individual sentences to determine sentiment, providing more detailed insights (Socher et. al. 2013).

**Aspect-Based Sentiment Analysis**: Analyzes sentiments related to specific aspects or features within the text, such as product reviews focusing on different attributes ( Pontiki et. al. 2016)

## 3. Traditional Approaches to Sentiment Analysis

### 3.1. Rule-Based Systems

Traditional sentiment analysis often relied on rule-based systems, which use predefined rules and linguistic resources to identify and classify sentiments.

**Lexicon-Based Approaches**: Lexicon-based methods use sentiment lexicons, which are lists of words associated with specific sentiments. Each word in the lexicon is assigned a sentiment score, and the overall sentiment of a text is determined by aggregating the scores of the words it contains. Examples include the AFINN and SentiWordNet lexicons (Baccianella et. al. 2010). These methods are straightforward but limited by the need for comprehensive lexicons and the inability to handle context-dependent meanings effectively.

**Pattern-Based Approaches**: Pattern-based methods involve creating rules based on syntactic and semantic patterns in the text. These rules can capture specific sentiment-bearing structures, such as phrases or negations, to determine sentiment (Pang et. al. 2008). While these methods can be more flexible than lexicon-based approaches, they often require extensive manual effort to develop and maintain.

### 3.2. Machine Learning Techniques

The advent of machine learning brought a shift from rule-based methods to approaches that learn from data.

**Feature Engineering**: In machine learning-based sentiment analysis, features such as term frequency-inverse document frequency (TF-IDF) or word embeddings are used to represent textual data. These features are then fed into classification algorithms to predict sentiment (Joachims et. al. 1998).. Feature engineering involves selecting and transforming raw text data into meaningful features to improve model performance.

**Classification Algorithms**: Various classification algorithms, such as Naive Bayes, Support Vector Machines (SVM), and Logistic Regression, are employed to classify text into sentiment categories (McCallum et. al. 1998). These algorithms learn patterns from labeled training data and apply them to new, unseen text. While effective, these methods often require careful tuning and feature selection to achieve optimal performance.

### 3.3. Limitations of Traditional Methods

Traditional approaches to sentiment analysis, while pioneering, have several limitations:

**Handling Complex Sentiments**: Traditional methods often struggle with complex sentiments, such as mixed or nuanced emotions, as they typically rely on binary or simple categorical classifications. These methods may not capture the full range of emotional expressions in the text.

**Generalizability Issues**: Rule-based and machine-learning approaches may have difficulty generalizing across different domains or languages. Rule-based systems depend heavily on domain-specific rules, while machine learning models may require extensive retraining to adapt to new contexts or languages.

## 4. Deep Learning Approaches in Sentiment Analysis

Deep learning has revolutionized sentiment analysis by enabling models to capture complex patterns and contextual information from text data. Neural networks, inspired by the human brain, consist of layers of interconnected nodes or neurons that process input data and learn features through training. Deep learning models, a subset of neural networks with many layers, have significantly advanced sentiment analysis by improving the ability to understand and generate text (LeCun et. al. 2015).

### 4.1. Recurrent Neural Networks (RNNs) and Long Short-Term Memory (LSTM) Networks

**Recurrent Neural Networks (RNNs)**: RNNs are designed to handle sequential data by maintaining a hidden state that captures information from previous steps in the sequence. This makes them suitable for text analysis, where the order of words and context is crucial. However, standard RNNs suffer from issues such as vanishing and exploding gradients, which can hinder their performance on long sequences (Bengio et. al. 1994).

**Long Short-Term Memory (LSTM) Networks**: LSTMs are a type of RNN designed to address the limitations of standard RNNs. They incorporate memory cells and gating mechanisms to regulate the flow of information, allowing the model to retain long-term dependencies and handle longer sequences more effectively (Hochreiter et. al. 1997). LSTMs have been widely used in sentiment analysis for their improved performance in capturing contextual information and handling complex sentence structures.

### 4.2. Convolutional Neural Networks (CNNs)

**Convolutional Neural Networks (CNNs)**: Originally developed for image processing, CNNs have been adapted for text analysis. CNNs use convolutional layers to automatically extract features from text data by applying filters that capture local patterns, such as phrases or n-grams (Kim et. al. 2014). This approach is particularly effective for identifying key sentiment-bearing phrases and patterns in text. CNNs are often combined with other models, such as RNNs or LSTMs, to enhance their performance in sentiment classification tasks.

### 4.3. Transformer Models

**BERT (Bidirectional Encoder Representations from Transformers)**: BERT is a transformer-based model that has significantly impacted sentiment analysis. It uses bidirectional attention to consider the context of each word in both directions (left-to-right and right-to-left), enabling it to capture nuanced meanings and relationships between words (Devlin et. al. 2019). BERT is pre-trained on a large corpus and can be fine-tuned for specific sentiment analysis tasks, leading to substantial improvements in accuracy and performance.

**GPT (Generative Pre-trained Transformer)**: GPT, developed by OpenAI, is another influential transformer-based model. Unlike BERT, GPT uses a unidirectional approach (left-to-right) for text generation and understanding. The model is pre-trained on diverse internet text and can be fine-tuned for various NLP tasks, including sentiment analysis (Radford et. al. 2018). GPT's ability to generate coherent and contextually relevant text has made it a valuable tool for understanding sentiment and generating responses.

### 4.4. Advantages and Challenges

**Improved Contextual Understanding**: Deep learning models, particularly transformers, offer significant advantages in understanding the context and nuances of sentiment expression. They can capture subtle emotional tones and complex sentence structures that traditional methods might miss.

**Computational Complexity**: Despite their advantages, deep learning models, especially transformers, require substantial computational resources and memory. Training these models involves processing large datasets and using powerful hardware, which can be a limitation for some applications (Vaswani et. al. 2017).

## 5. Large Language Models (LLMs) and Their Impact

*Large Language Models (LLMs) have significantly transformed the field of sentiment analysis through their advanced capabilities in understanding and generating natural language. This section explores the evolution of LLMs, their applications, and their impact on sentiment analysis.*

### 5.1. Evolution of LLMs

*Large Language Models have evolved from early statistical models to sophisticated deep learning architectures. Initially, sentiment analysis was performed using simpler models such as bag-of-words and n-grams, but these approaches had limitations in capturing contextual information and semantic nuances (Manning et. al. 2008)].*

*The advent of neural network-based models brought significant improvements, with embeddings like Word2Vec and GloVe offering richer representations of words by capturing semantic relationships (Mikolov et. al. 2013). However, the real breakthrough came with the introduction of transformer-based models, which leverage self-attention mechanisms to process and understand the text more nuancedly* (Vaswani et. al. 2017).

### 5.2. Pre-trained Models and Fine-Tuning

*Pre-trained LLMs, such as BERT and GPT, have set new benchmarks in various NLP tasks, including sentiment analysis. These models are trained on vast amounts of text data and learn to generate contextual representations of words and phrases.*

> ***BERT (Bidirectional Encoder Representations from Transformers)**: BERT, introduced by Devlin et al., uses a bidirectional approach to capture context from both directions (left-to-right and right-to-left). This bidirectional training allows BERT to understand the meaning of words based on their surrounding context, leading to more accurate sentiment classification* (Devlin et. al. 2019). *BERT is pre-trained on tasks such as masked language modeling and next-sentence prediction, and it can be fine-tuned for specific sentiment analysis tasks with relatively small datasets.*

***GPT (Generative Pre-trained Transformer)***: GPT, developed by OpenAI, employs a unidirectional approach and is known for its ability to generate coherent and contextually relevant text. GPT's pre-training involves predicting the next word in a sequence, which enables it to capture syntactic and semantic patterns in text. Fine-tuning GPT on sentiment analysis tasks can enhance its performance in understanding and generating sentiment-laden text.

### 5.3. Multimodal Sentiment Analysis

Recent advancements in LLMs have also led to the exploration of multimodal sentiment analysis, where textual data is integrated with other modalities such as images or audio. This approach aims to capture sentiment more comprehensively by considering multiple sources of information (Radford et. al. 2021).

***Integration of Text with Other Modalities***: Models like CLIP (Contrastive Language-Image Pretraining) combine text and image data to improve sentiment analysis by leveraging visual context along with textual information. This integration helps in understanding sentiment in a more nuanced way, especially in contexts where visual elements contribute significantly to the overall sentiment (Ramesh et. al. 2021).

## 6. Challenges in Sentiment Analysis

Despite the advancements in sentiment analysis, several challenges persist that affect the accuracy and effectiveness of sentiment analysis systems. This section discusses key challenges and potential solutions in sentiment analysis.

### 6.1. Sentiment Detection in Bilingual and Multilingual Texts

***Cross-Lingual Models***: Sentiment analysis across different languages presents unique challenges due to variations in linguistic structures, cultural contexts, and sentiment expressions. Cross-lingual models aim to address these issues by leveraging shared semantic spaces or aligning embeddings across languages. Techniques such as multilingual embeddings and transfer learning have been employed to build models that can understand and analyze sentiments in multiple languages.

***Domain-Specific Adaptations***: Sentiment expression can vary significantly across domains. For example, sentiment related to financial markets may be expressed differently than sentiment in social media. Domain adaptation techniques involve fine-tuning models on domain-specific data to improve their

*performance in particular contexts. This adaptation is crucial for achieving accurate sentiment analysis in specialized fields (Bolukbasi et. al. 2016).*

***6.2. Detection of Sarcasm and Irony***

***Challenges and Solutions****: Sarcasm and irony pose significant challenges for sentiment analysis because they often invert the literal meaning of words. Detecting these forms of sentiment requires models to understand context, tone, and sometimes cultural nuances. Approaches such as using context-aware models (e.g., transformers) and incorporating external knowledge sources have shown promise in improving sarcasm detection. Additionally, annotated datasets with sarcastic and ironic examples can help train more robust models.*

***6.3. Bias and Fairness in Sentiment Analysis***

***Identifying and Mitigating Biases****: Sentiment analysis models can inherit and perpetuate biases present in training data. These biases may relate to demographic factors such as gender, race, or socioeconomic status, leading to unfair or inaccurate sentiment classification. Techniques to identify and mitigate biases include diverse data collection, fairness-aware training algorithms, and bias detection frameworks. Ensuring that models are trained on balanced and representative datasets is crucial for minimizing bias (Gupta et. al. 2024).*

***Ethical Considerations****: The ethical implications of sentiment analysis extend beyond bias. Issues such as privacy, data security, and the potential misuse of sentiment data must be considered. Implementing ethical guidelines and ensuring transparency in how sentiment data is used can help address these concerns. Moreover, user consent and data anonymization practices should be prioritized to protect individual privacy.*

***7. Future Trends and Research Directions***

*The field of sentiment analysis is continuously evolving, driven by advancements in technology and the growing demand for sophisticated text understanding. This section explores the emerging trends and future research directions that are likely to shape the landscape of sentiment analysis in the coming years.*

***7.1. Advances in Deep Learning Architectures***

***New Architectures and Techniques****: The development of novel deep learning architectures promises to enhance sentiment analysis capabilities. Techniques such as graph neural networks (GNNs) and attention mechanisms are being explored for their potential to capture complex relationships and contextual*

*information in text. These advancements could lead to more accurate and nuanced sentiment analysis by better understanding the interplay between different elements of a text (Wu et. al. 2021).*

***Enhanced Transfer Learning***: *Transfer learning continues to be a significant area of research, with a focus on improving the effectiveness of pre-trained models for sentiment analysis. Advances in few-shot learning and zero-shot learning techniques could enable models to perform sentiment analysis with minimal labeled data, making them more adaptable to new domains and languages (Brown et. al. 2020).*

### 7.2. Enhanced Handling of Context and Nuance

***Contextualized Sentiment Analysis***: *Future research is likely to focus on improving the ability of sentiment analysis models to handle nuanced contexts. This includes a better understanding of sarcasm, irony, and subtle emotional cues. Innovations in context-aware models and multi-turn dialogue systems may provide deeper insights into the sentiment expressed in complex or conversational texts (Zhang et. al. 2020).*

***Emotion Detection***: *Beyond simple sentiment classification, there is growing interest in detecting specific emotions such as joy, anger, or sadness. This requires models that can distinguish between different emotional states and understand their implications in various contexts. Advances in emotion-aware sentiment analysis could provide more detailed insights into user sentiments and reactions (Chutia et. al. 2024).*

### 7.3. Future Research Directions

***Cross-Linguistic and Cross-Cultural Analysis***: *Future research will likely explore cross-linguistic and cross-cultural aspects of sentiment analysis to improve models' effectiveness across diverse languages and cultures. This includes developing techniques for handling multilingual data and understanding cultural varionations in sentiment expressions (Miah et. al. 2024)*

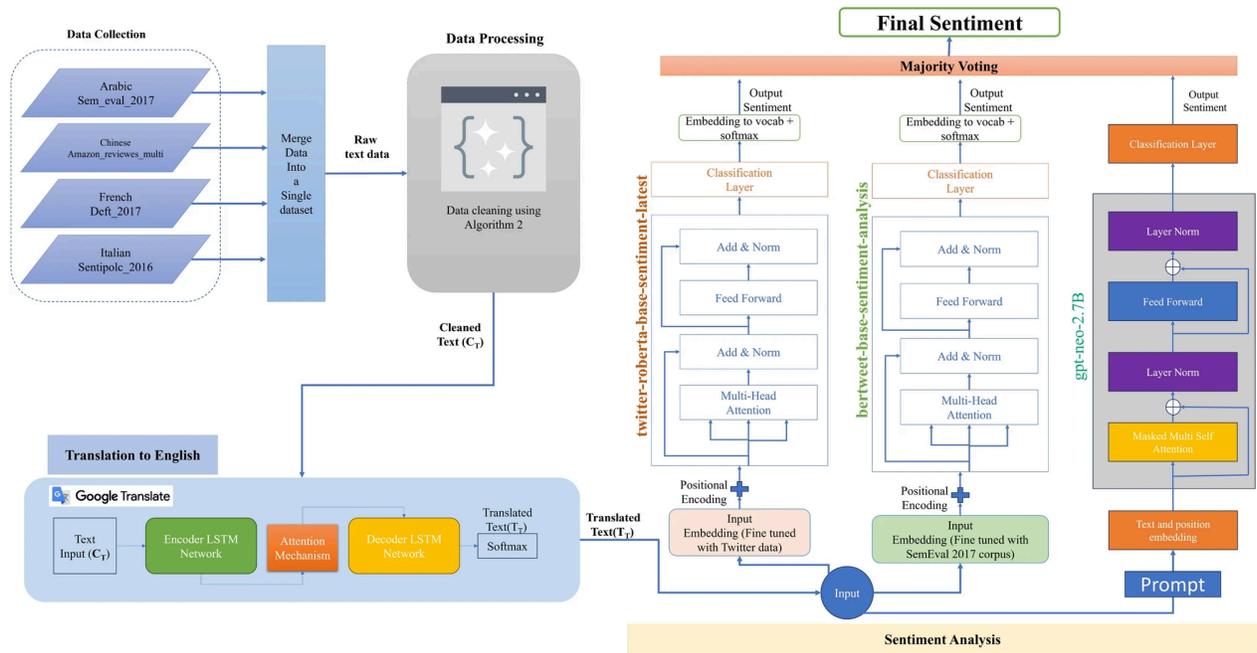

*Figure 1: Overview of a multimodal approach to cross-lingual sentiment analysis (Miah et. al. 2024)*

**Explainable AI**: Enhancing the interpretability and transparency of sentiment analysis models is an important research direction. Explainable AI techniques (Uddin et. al. 2024) can help users understand how models arrive at their conclusions, increasing trust and accountability in sentiment analysis applications.

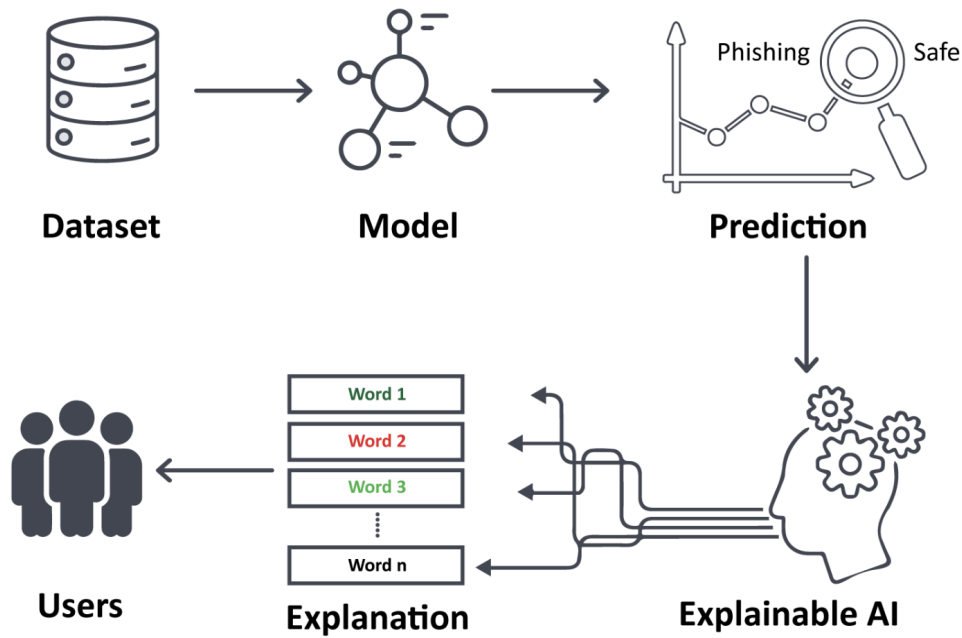

*Figure 2: Overview of an Explainable Transformer-based Model (Uddin et. al. 2024)*

**Human-AI Collaboration**: The integration of human expertise with AI-driven sentiment analysis can lead to more accurate and contextually relevant results. Future research may focus on developing systems that facilitate effective collaboration between human analysts and AI models, combining the strengths of both to improve sentiment analysis outcomes.